\documentclass{osa-article}

\journal{osajournal}

\begin{document}

\title{Uncalibrated Deflectometry with a Mobile Device  \\ on Extended Specular Surfaces}

\author{Florian Willomitzer,\authormark{1,*} Chia-Kai Yeh,\authormark{1} Vikas Gupta,\authormark{1} William Spies,\authormark{1} Florian Schiffers,\authormark{1} Marc Walton,\authormark{2} Oliver Cossairt\authormark{1}}

\address{\authormark{1}Department of Computer Science, Northwestern University, Evanston, IL 60208 \\
\authormark{2}Center for Scientific Studies in the Arts, Northwestern University, Evanston, IL 60208}

\email{\authormark{*}florian.willomitzer@northwestern.edu} 



\begin{abstract}

We introduce a system and methods for the three-dimensional measurement of extended specular surfaces with high surface normal variations. Our system consists only of a mobile hand held device and exploits screen and front camera for Deflectometry-based surface measurements. We demonstrate high quality measurements without the need for an offline calibration procedure. In addition, we develop a multi-view technique to compensate for the small screen of a mobile device so that large surfaces can be densely reconstructed in their entirety. This work is a first step towards developing a self-calibrating Deflectometry procedure capable of taking 3D surface measurements of specular objects in the wild and accessible to users with little to no technical imaging experience.
\end{abstract}

\section{Introduction}

\begin{figure}[b]
\centering \includegraphics[width= 0.7\columnwidth]{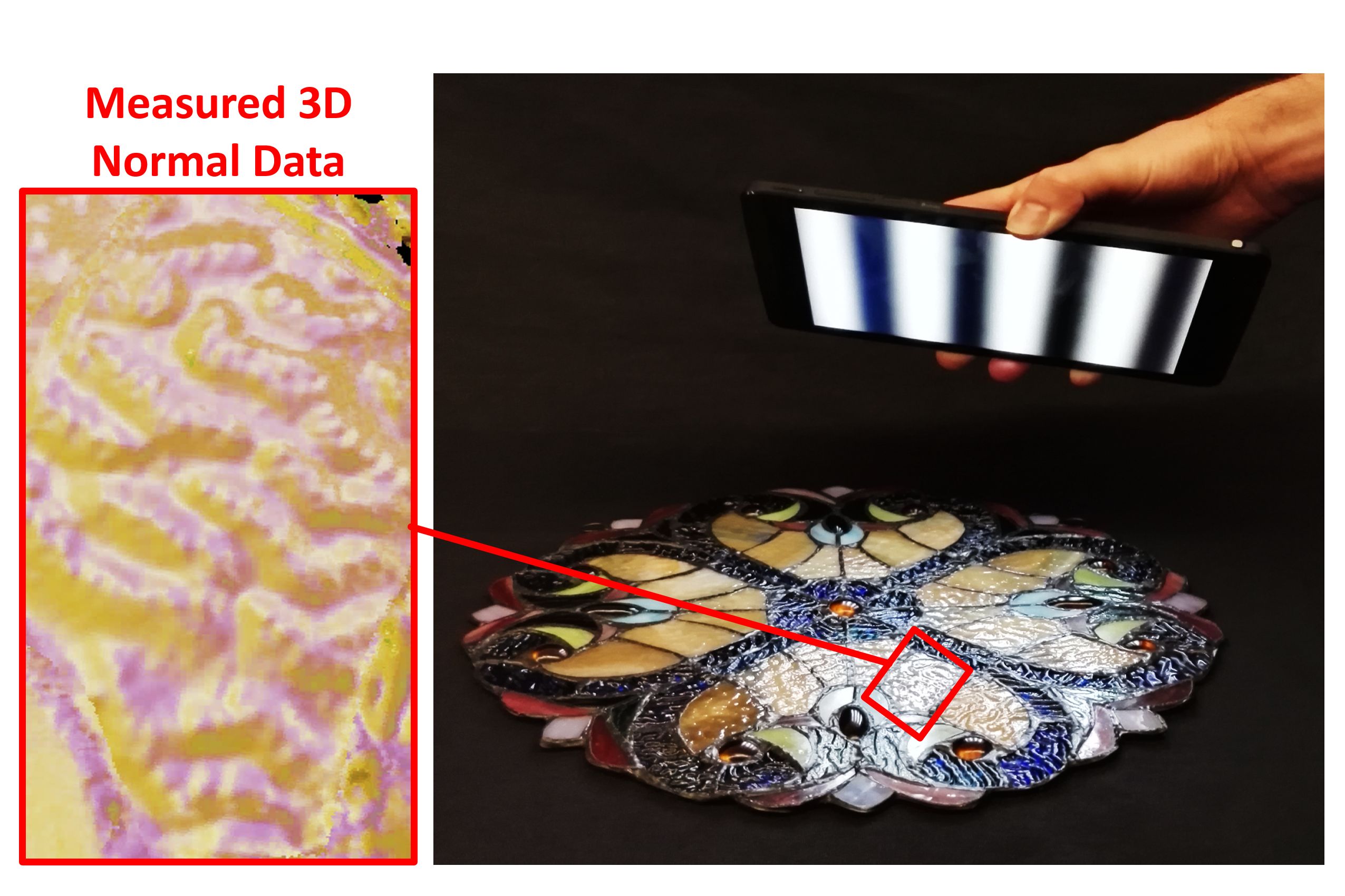}
\caption{Hand held measurement of a stained glass painting with a mobile device. The reflections of the screen are visible on parts of the glass surface and reveal it's three-dimensional structure. The measurement result (normal map) is displayed in the zoomed inset.}
\label{fig:OE_Title}
\end{figure}

Three-dimensional (3D) imaging techniques are now omnipresent in a multitude of scientific and commercial disciplines. Industrial 3D~inspection, medical 3D~imaging as well as 3D~documentation and analysis of art or cultural heritage are only a few examples of the broad range of applications. The great popularity of 3D imaging is no coincidence: Compared	to	a	simple	2D	image, a	three-dimensional	object	representation	is invariant to	object translation and	rotation,	as	well	as	variations	in	surface	texture or	external	illumination	conditions.

Unfortunately, these benefits do not come without a price: 3D image acquisition is not as easy and straightforward as taking a 2D snapshot with a mobile phone camera. As a first step, one must pick the appropriate 3D imaging technique for the object to be measured. This decision is strongly dependent on the object's microscopic surface structure, which can be roughly divided into two categories: \textit{(diffuse) scattering} and \textit{specular}.

Diffusely scattering surfaces are commonly measured by projecting a temporally or spatially structured light beam onto the object and evaluating the back-scattered signal. `Time-of-Flight'~\cite{doi:10.1117/12.287751} or Active Triangulation (`Structured Light')~\cite{Willomitzer:17} are prominent examples. Another well known principle is `Photometric Stereo'~\cite{doi:10.1117/12.7972479}, where the object surface is sequentially flood illuminated with `point' light sources from different angles.

Unfortunately, the application of these principles to \textit{specular surfaces} yields only limited success. The reason for this is simple: specular reflections from a point light source scarcely find their way back into the camera objective. Depending on the distribution of surface normals with respect to light source and camera position, only a few sparse (and probably overexposed) `specular spots' may be visible in the camera image. 
The same problem is also common for interferometric instruments, which can accurately measure smoothly varying specular surfaces with high precision, but fail  for specular surfaces with a large angular distribution of surface normals.

\begin{figure}[b]
\centering \includegraphics[width= 0.7\columnwidth]{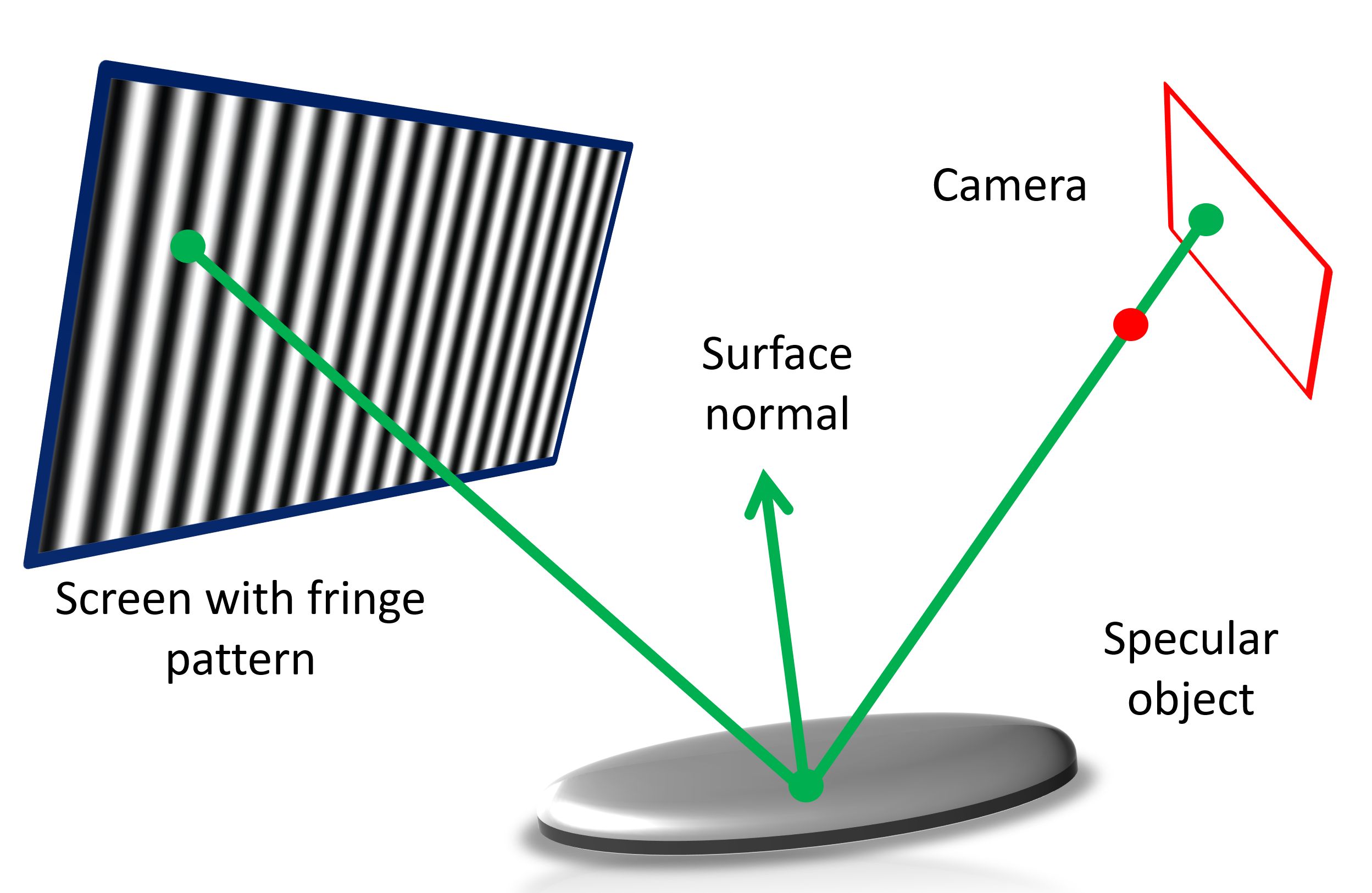}
\caption{Basic principle of `Phase Measuring Deflectometry' (PMD): A screen with a fringe pattern is observed over the reflective surface of an object. The normal map of the object surface can be calculated from the deformation of the fringe pattern in the camera image.}
\label{fig:OE_PMDBasic}
\end{figure} 

A straightforward solution to this problem is to extend the angular support of the illumination sources. This is the basic principle behind \textit{`Deflectometry'}~\cite{gerd1999, doi:10.1117/12.545704}, where a patterned screen replaces `point-like' light source (see Fig~\ref{fig:OE_Title} or Fig~\ref{fig:OE_PMDBasic}). This screen can be self illuminated (TV Monitor) or printed. In Deflectometry systems, the screen and camera face the object, which means that the camera observes the \textit{specular reflection} of the screen over the object surface. The observed pattern in the camera image is a deformed version of the image on the screen, where the deformation depends on the surface normal distribution of the object surface (Fig~\ref{fig:OE_PMDBasic}). From this deformation, the normal vectors of the surface can be calculated. To calculate a normal vector at each camera pixel, correspondence between camera pixels and projector pixels must be determined. A common technique to achieve this is with the phase-shifting of sinusoidal fringes. The resulting \textit{`Phase-Measuring Deflectometry'} (PMD)~\cite{gerd1999, doi:10.1117/12.545704} has established itself as powerful technique that is used with great success in industrial applications, e.g. to test the quality of optical components or to detect defects on metallic parts like car bodies. Given a proper calibration, PMD reaches precisions close to interferometric methods~\cite{doi:10.1117/12.957465, Olesch2012DeflectometricSF, doi:10.1117/12.2020578}. \\

The work introduced in this paper was motivated by a specific and challenging application of PMD: The \textit{3D measurement and analysis of stained glass paintings}, which can be found in larger glass artworks, church windows, or glass reliefs. Such glass paintings can be composed of hundreds or thousands small colorful glass pieces that are connected with a metal frame. The shape of these small glass pieces is not necessary flat! Over the centuries, several glass manufacturers developed a multitude of techniques to imprint unique three-dimensional structures to the glass surface that reflect and diffract light in a very distinct way. These unique 3D structures can be exploited to match the small glass pieces in a stained glass painting to the individual manufacturers and to trace the circulation of stained glass and the respective historical influence of the manufacturer around the globe. The latter is of significant interest for the cultural heritage community. \\

The task of digitizing  3D surface types similar to the surfaces of stained glass paintings leads to several fundamental and technical challenges of  great scientific interest for the 3D metrologist:

\begin{itemize}
\item The objects (e.g. church windows) are \textbf{large} and usually \textbf{not portable}. This makes it nearly impossible to transport them to a controlled lab environment for surface measurement.

\item The surfaces to be measured contain \textbf{high spatial frequencies} and thus a \textbf{large variation of surface normals}. This requires a high spatial resolution and angular coverage of the 3D surface measurement device.

\item The \textbf{backside} of the objects is largely \textbf{inaccessible}, which rules out all 'spectroscopy-like' methods to characterize the 3D surface. Moreover, some objects can be highly opaque.

\end{itemize}

All of the above challenges can be addressed by using a PMD measurement technique. However, most PMD setups are bulky and cannot be applied 'in the wild'. This bulkiness is mainly caused by the large screens used, which are intentionally chosen to provide a large angular coverage and enable measurement over a large range of surface normals. 

Our solution to this problem is to use \textit{mobile devices} (smart phone, tablet) for PMD measurements, i.e. \textit{using the screen to display the patterns} and the \textit{front-facing camera to image the object surface}. Since the screen size of mobile devices is limited, only a small angular range of surface normals can be measured in any single view. We propose to overcome this limitation using \textit{feature-based registration}, applied to \textit{multiple views acquired from different viewing angles}. The features are extracted directly from the glass surface so that external markers or fiducials are not necessary.\\

We seek to develop a 3D measurement tool that can be adopted by a broad audience of users with little to no technical expertise. In particular, we wish to provide 3D surface measurement capability to museum conservators and tourists, who may require extreme ease of use and minimal computational requirements. 

To make our system widely accessible, we are working on a platform that allows for a server-side evaluation of the 3D data \cite{Science, Smith}. The mobile device is used only to \textit{capture} images and to \textit{display} the evaluated data. 

\newpage

\section{Related Work} \label{sec:RelWork}
`Phase Measuring Deflectometry' (PMD) is just one of a number of techniques that have been introduced to measure the 3D surface of specular objects. Most of these techniques are closely related to PMD, but differ in the mechanism used to establish correspondence between the screen and camera. In principle \textit{any known pattern} can be used in place of sinusoidal fringes. Furthermore, the pattern does not even have to be self illuminated, and will be effective in estimating surface shape as long as there is some prior knowledge of the pattern.

The authors of~\cite{1238401} used the reflection of color coded circles observed by multiple cameras (which also resolves the bas-relief ambiguity). In other works, the authors utilized self illuminated screens with patterns such as stripes~\cite{TARINI2005233}, multiple lines~\cite{5206624}, or even a light field created from two stacked LED screens~\cite{7492867}. `Screenless' methods, such as \cite{7335465, 6751198} analyze environment illumination or track prominent features (e.g. straight lines) in the environment to obtain information about the slope of specular surfaces.

Each of the above mentioned techniques comes with benefits and drawbacks. For example, some of the techniques that use a static pattern instead of a phase shifted sinusoid are capable of `single-shot' acquisition ~\cite{LiuSS}. However, this comes at the cost of lateral resolution and/or restricts the surface frequencies that can be measured. 

Self-calibrating photometric stereo techniques use known reflectance maps of object surfaces to measure their 3D structure \cite{Ikeuchi:1992:DSO:136740.136778, Tunwattanapong:2013:ARS:2461912.2461944}. Such approaches are especially beneficial for partially specular surfaces, but fail when the surface is too shiny. Other techniques exploit sparse specular reflections produced by photometric stereo measurements for 3D surface reconstruction or refinement ~\cite{Chen:2006:MS:1153171.1153665, 3866, 54736}.

Mobile versions of Deflectometry have been demonstrated as well. The authors of~\cite{Roettinger2011Deflectometric}, built a custom Deflectometry device compact enough to be used inside diamond turning machines. The work most closely related to ours is  \cite{10.1111:cgf.12719, doi:10.1117/1.OE.54.2.025111}. The authors also  exploit the LCD screen and front camera of a smartphone or tablet to perform Deflectometry measurements. However, both of these works demonstrate results with limited field of view and normal coverage. 3D surface measurement for objects with high frequency surface information is not addressed in these papers. Therefore, these systems are not suited for the 3D surface measurement of \textit{extended specular surfaces} with \textit{large normal variations}, such as stained glass paintings.

\section{Towards Self-Calibrating Phase-Measuring Deflectometry}

This section describes the image acquisition and processing steps that enable uncalibrated 3D Deflectometry measurements with mobile devices. We demonstrate a set of qualitative surface measurements that are sufficient to identify and compare characteristic surface structures, e.g. in stained glass surfaces. Eventually, we apply our method to other (partially) specular surface types, like oil paintings or technical parts.

\subsection{Setup and Image Acquisition Process}

Our implementation consists of the following components:

\begin{itemize}
\item A consumer tablet that serves as measurement device (for the results shown in this paper we used a NVIDIA Shield K1 or an Apple iPad Pro 10.5")
\item A controlling device (in our case a laptop or a second tablet) that is used to start and monitor the measurement and to display the final result
\item A server that performs the 3D evaluation and hosts the respective code
\item An application that handles the image acquisition process and manages the data transfer between server, measurement device and controlling device
\end{itemize}

Besides the aspect of higher performance, evaluating the data on a server has other important benefits: Code changes (updates) can be directly made on the server without the need for the user to install a new version of the measurement App. Moreover, the server can store the evaluated data and work as a database, e.g. for the identification of stained glass pieces or similar `fingerprint-applications'.

During image acquisition, the tablet displays phase-shifted sinusoidal patterns and observes the object  with its front camera (see Fig.~\ref{fig:OE_Title}). The tablet is positioned approximately 200 mm over the object surface. PMD is a \textit{multi-shot principle}, meaning that \textit{a sequence of temporally acquired images} has to be used to calculate \textit{one 3D image}. During the measurement, the display projects four $90^\circ$-phase-shifted versions of a sinusoid in horizontal and vertical direction, respectively. Different frequencies of the sinusoid can optionally be used as well. The position of the tablet relative to the object has to remain fixed during the whole acquisition process. Depending on the speed of projection and image acquisition, this can be a hard task for the inexperienced user, if a handheld measurement is desired. For an optimal measurement result, the tablet should be fixed with a respective mount. We discuss extension of our system towards a \textit{free-hand guided single-shot principle} in section~\ref{sec:ConcOut}.

The front-camera objectives of mobile device cameras commonly have a short focal length which results in a large field of view. Unfortunately, this large field of view cannot be exploited in its entity by our system. A valid PMD measurement can only be taken at image pixels that \textit{observe a display pixel} over the specular surface. This is because the device cannot be held closer to the object surface than the minimum possible focus distance and the LCD screen has a limited angular coverage. As a result, the number of pixels that produce valid measurements can be as small as 25\% of the imaging field of view. Figure~\ref{fig:GradMaps}a illustrates this problem with a planar object (plexiglass plate), which is placed at the minimum possible distance to the tablet. The plate is larger than the field of view of the camera, nevertheless, the pattern can be only observed in a small portion of the field.
\subsection{Evaluation, Results and Discussion}

In the following, we evaluate the surface normal map of stained glass test tiles which have been provided to us by the Kokomo opalescent glass company~\cite{kokomo}. A photo of the test tiles is shown in Fig.~\ref{fig:TileTestSet}. The tiles have an approximately square shape with edge length of about 50 mm. Surface structure complexity and angular distribution of surface normals increase from a) to d) in Fig.~\ref{fig:TileTestSet}.  Eventually, we show and discuss a `multi-view measurement' of an extended stained glass painting.\\

\subsubsection{Single-View Measurement} \label{sec:SingleView}

\begin{figure}[t!]
\centering \includegraphics[width= 0.8\columnwidth]{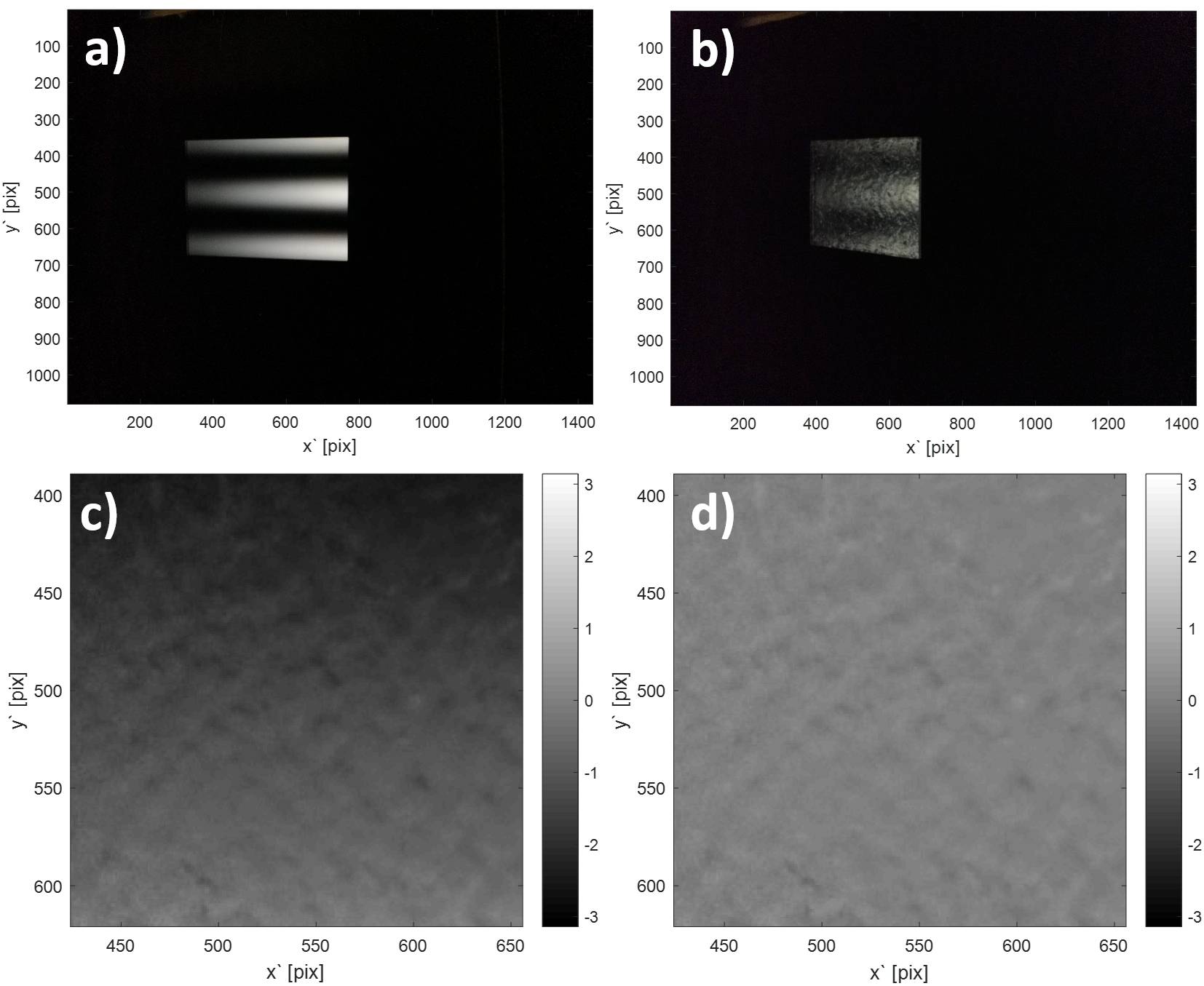}
\caption{a) Horizontal fringes within the effective measurement field for a planar object (plexiglass plate). b) Horizontal fringe pattern, reflected from the surface of test tile `33~KDR'. c) Calculated  phase map. d) Calculated gradient map after high pass filtering.}
\label{fig:GradMaps}
\end{figure}

\begin{figure}[p]
\centering \includegraphics[width= 0.65 \columnwidth]{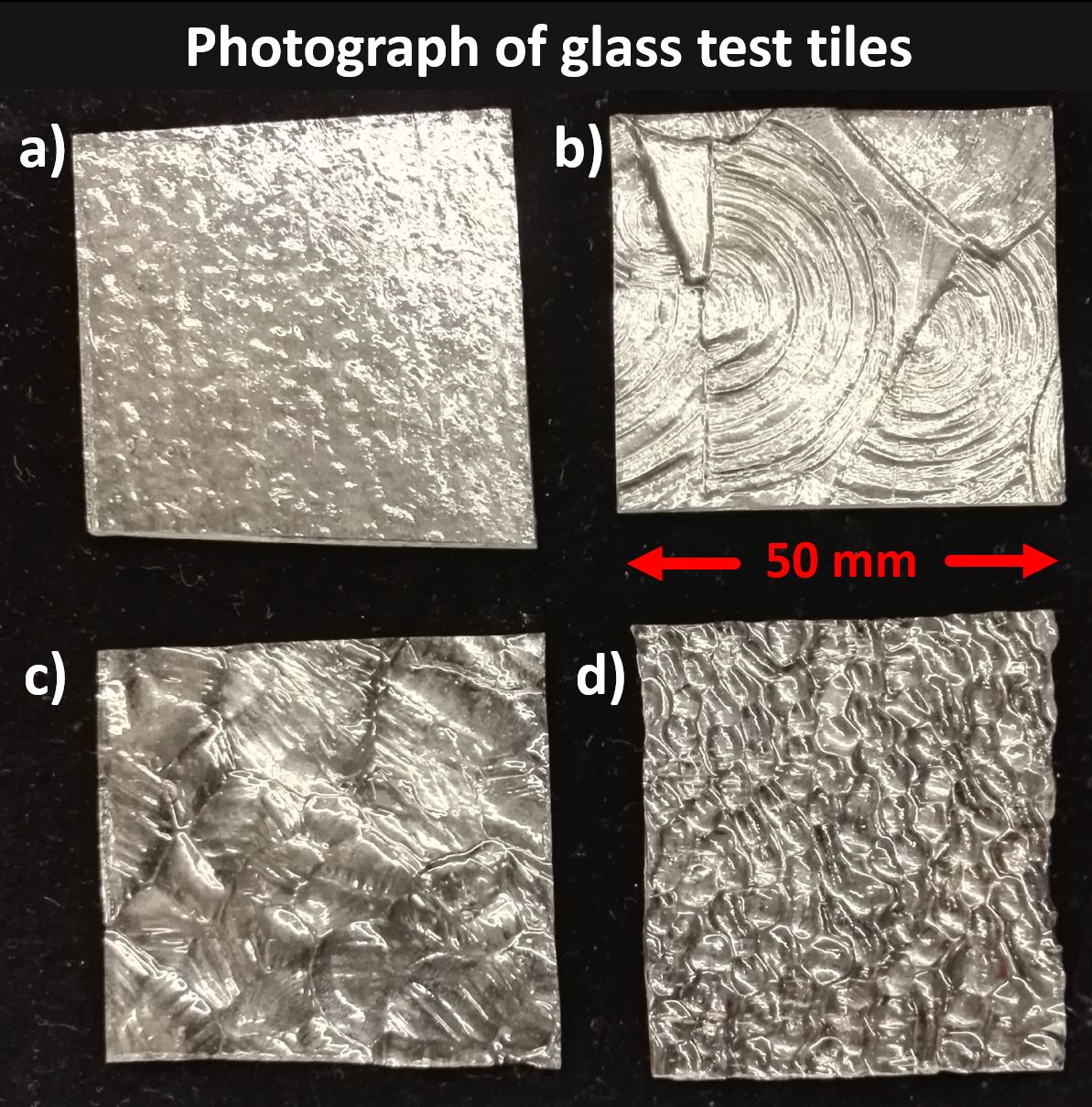}
\caption{Set of measured stained glass test tiles. Surface structure complexity and angular distribution of surface normals increase from a to d. Glass tile nomenclatures: a) `33KDR' b) `33 RON' c) `33WAV' d) `33TIP'.}
\label{fig:TileTestSet}

 \vspace{5mm}

\centering \includegraphics[width= 0.65\columnwidth]{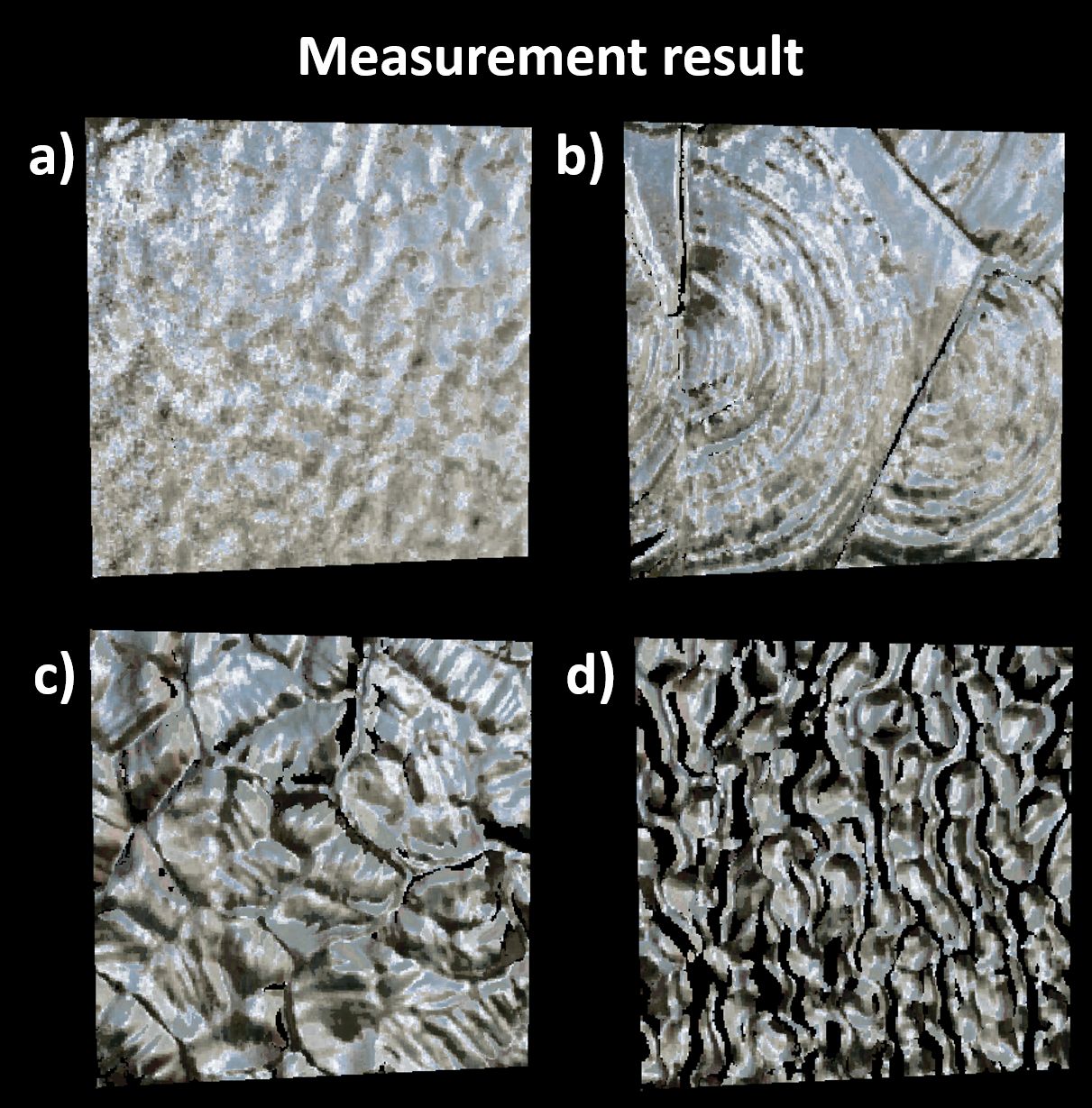}
\caption{ Surface normal maps, calculated from the measurements of the test tile set from~Fig.~\ref{fig:TileTestSet}. a) `33KDR' b) `33 RON' c) `33WAV' d) `33TIP'.}
\label{fig:NormalMap4Tiles}
\end{figure}

Most of the tiles in our test set display a size and surface normal distribution small enough to be evaluated from one single view. For illustration purposes, we explain the measurement process for the`easiest' test tile (Fig.~\ref{fig:TileTestSet}a). The tile is placed at a position in the field  of view, where the reflected display can be observed (see Fig.~\ref{fig:GradMaps}b). The intensity in each image pixel $(x',y')$ can be expressed as 

\begin{equation}
\label{eq:PS1}
I(x',y') = A(x',y') + B(x',y') \cdot \cos(\phi(x',y'))~.
\end{equation}

Eq.~(\ref{eq:PS1}) contains \textit{three unknowns per pixel}: The (desired) phase $\phi(x',y')$ of the sinusoidal pattern, that correlates display pixels with image pixels, but also $A(x',y')$ and $B(x',y')$, which contain information about the unknown bias illumination and object reflectivity. This means that at least \textit{three equations} are required to calculate $\phi(x',y')$. For each pattern direction, these equations are taken from the four\footnote{The \textit{four} phase-shift algorithm is very simple and, in addition, insensitive to second order nonlinearities} acquired phase-shift images, where the intensity in each image pixel for the $m$-th phaseshift is

\begin{equation}
\label{eq:PS2}
I_m(x',y') = A(x',y') + B(x',y') \cdot \cos(\phi(x',y') - \phi_m)~, 
\end{equation}

with

\begin{equation}
\label{eq:PS3}
\phi_m = (m-1) \frac{\pi}{2} ~.
\end{equation}

Finally $\phi(x',y')$ can be evaluated by

\begin{equation}
\label{eq:PS4}
\phi(x',y') = \arctan{\frac{I_2(x',y') - I_4(x',y') }{I_1(x',y') - I_3(x',y')}}
\end{equation}

This has to be done for each pattern direction, leading to phase maps $\phi_x(x',y')$ and $\phi_y(x',y')$ for the horizontal and vertical fringe direction respectively. The  phase map for the horizontal fringe direction is shown in Fig.~\ref{fig:GradMaps}c (image cropped for better visualization). This phase map is equivalent to the surface gradient in the vertical direction \textit{plus} a low frequency phase offset  that is dependent on the relative position between camera and object, and any distortion present in the camera objective \cite{doi:10.1117/12.545704, Olesch2012DeflectometricSF}. In conventional PMD setups, this offset is removed by employing a calibration process whereby the phase map is first measured for a planar mirror, then subtracted from the measured phase. For the desired qualitative measurements of objects like stained glass artworks, we can avoid this step and the respective `phase calibration' by exploiting a-priori knowledge about our objects, namely that their overall shape is \textit{always flat!} In this case, the unknown  phase offset can be removed by simply high pass filtering the unwrapped phase map. The high pass filtered phase maps $\tilde{\phi_x}$ and $\tilde{\phi_y}$ are then equivalent to the surface gradient maps  in x- and y- direction.  It should be noted that the filtering operation also compensates for the nonlinear photometric responses of the display and camera, avoiding an additional calibration procedure. Moreover, the assumption of a flat object resolves the depth-normal ambiguity of Deflectometry  measurements which typically requires 2 cameras to resolve~\cite{doi:10.1117/12.545704} . The resulting horizontal gradient map is displayed in Fig.~\ref{fig:GradMaps}d. The results were calculated using the pattern frequency $\nu=1$, corresponding to one sinusoidal period displayed over the entire width of the screen. Measurements with frequencies $\nu>1$ and subsequent \textit{multi-frequency phase unwrapping} can be performed as well. These measurements are not shown in this paper because the approach does not significantly improve the performance of 3D surface measurements and needs more time to acquire the necessary fringe-images.   

\begin{figure}[b]
\centering \includegraphics[width= 0.7 \columnwidth]{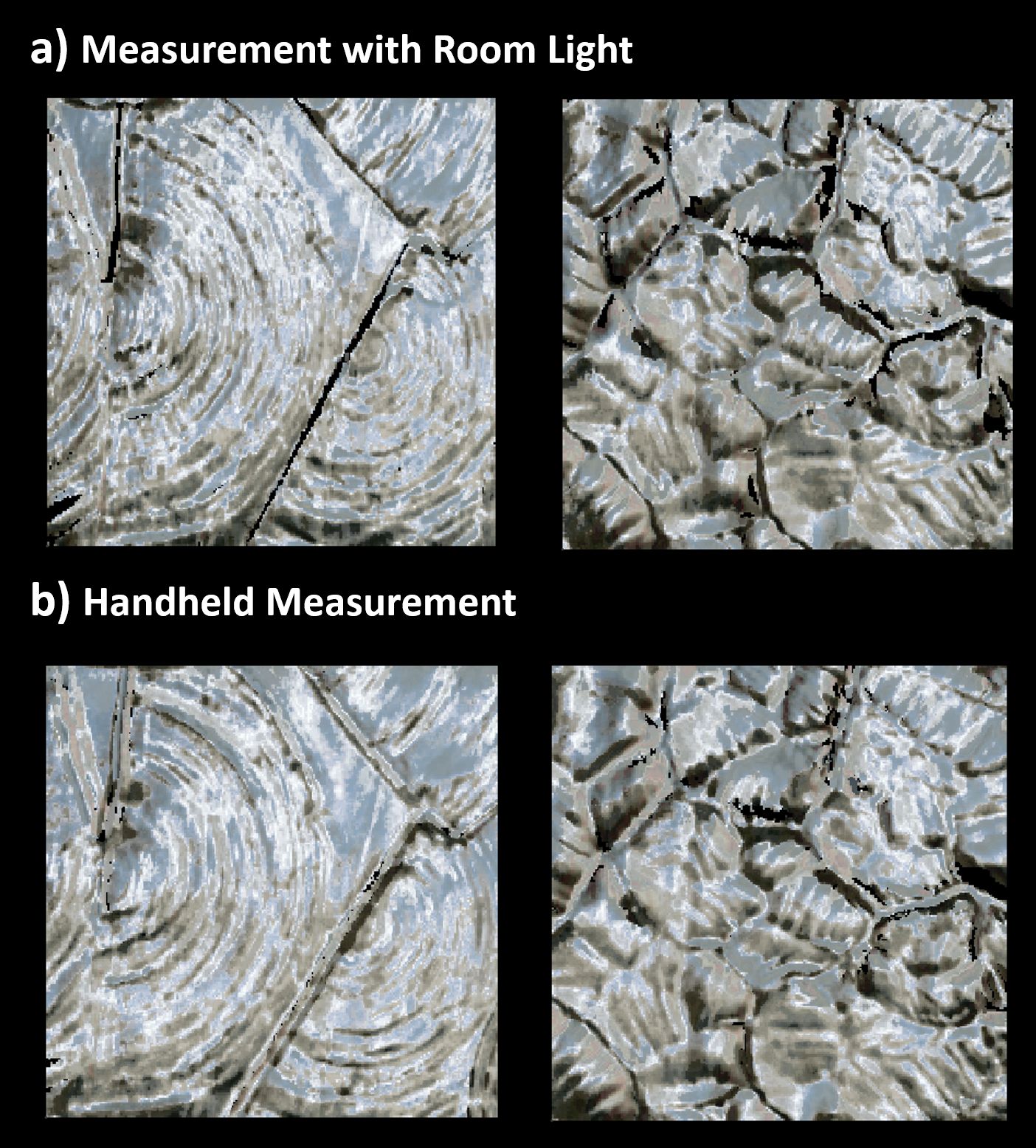}
\caption{Normal map measurement results for the test tiles `33 RON' and `33WAV'. a)~Measurement under room light. b) Handheld measurement.}
\label{fig:RoomHand}
\end{figure}

The surface normal can be computed directly from the estimated phase maps via

\begin{equation}
\label{eq:Normal}
\vec{n} = \frac{1}{\sqrt{\tilde{\phi_x}^2 + \tilde{\phi_y}^2 +1}} \cdot \left(\begin{array}{c}\tilde{\phi_x}\\\tilde{\phi_y}\\-1\end{array}\right)~,
\end{equation}

where $\tilde{\phi_x}$ and $\tilde{\phi_y}$ denote the gradient for the horizontal and vertical direction, respectively. Figure~\ref{fig:NormalMap4Tiles} shows the calculated normal maps of all 4 tiles. The 4 normal maps are shaded with a specular finish and slightly tilted for visualization purposes. It can be seen that the characteristic surface structures are well resolved. The black spots in the normal maps are produced by surface points where the surface normal resulted in no measured signal, i.e. the camera was not able to see the display.

To test the robustness of our qualitative measurement results against different environmental conditions, we additionally acquired measurements for two of the four tiles with ambient room lighting and with performing a hand-held measurement without mounting the device. The results are shown in Fig.~\ref{fig:RoomHand}.

The measurement taken with ambient room lighting (Fig.~\ref{fig:RoomHand}a) shows no significant degradation in performance. This is understandable because the brightness of the room light was moderate and the SNR was not reduced significantly. Under these conditions, the four-phaseshift algorithm effectively compensates for bias illumination. For the free-hand guided measurement, motion artifacts in the evaluated phase map are expected. These artifacts can be seen at the slightly blurred edges in Fig.~\ref{fig:RoomHand}b. The fact that the visible artifacts occur `only' at edges is a consequence of the low frequency $\nu =1$ used to acquire these measurements. Higher frequencies would result in more prominent artifacts, as for example commonly observed in triangulation-based fringe projection. \\

\subsubsection{Multi-view measurement}  \label{sec:MultView}

\begin{figure}[b]
\centering \includegraphics[width= 0.7 \columnwidth]{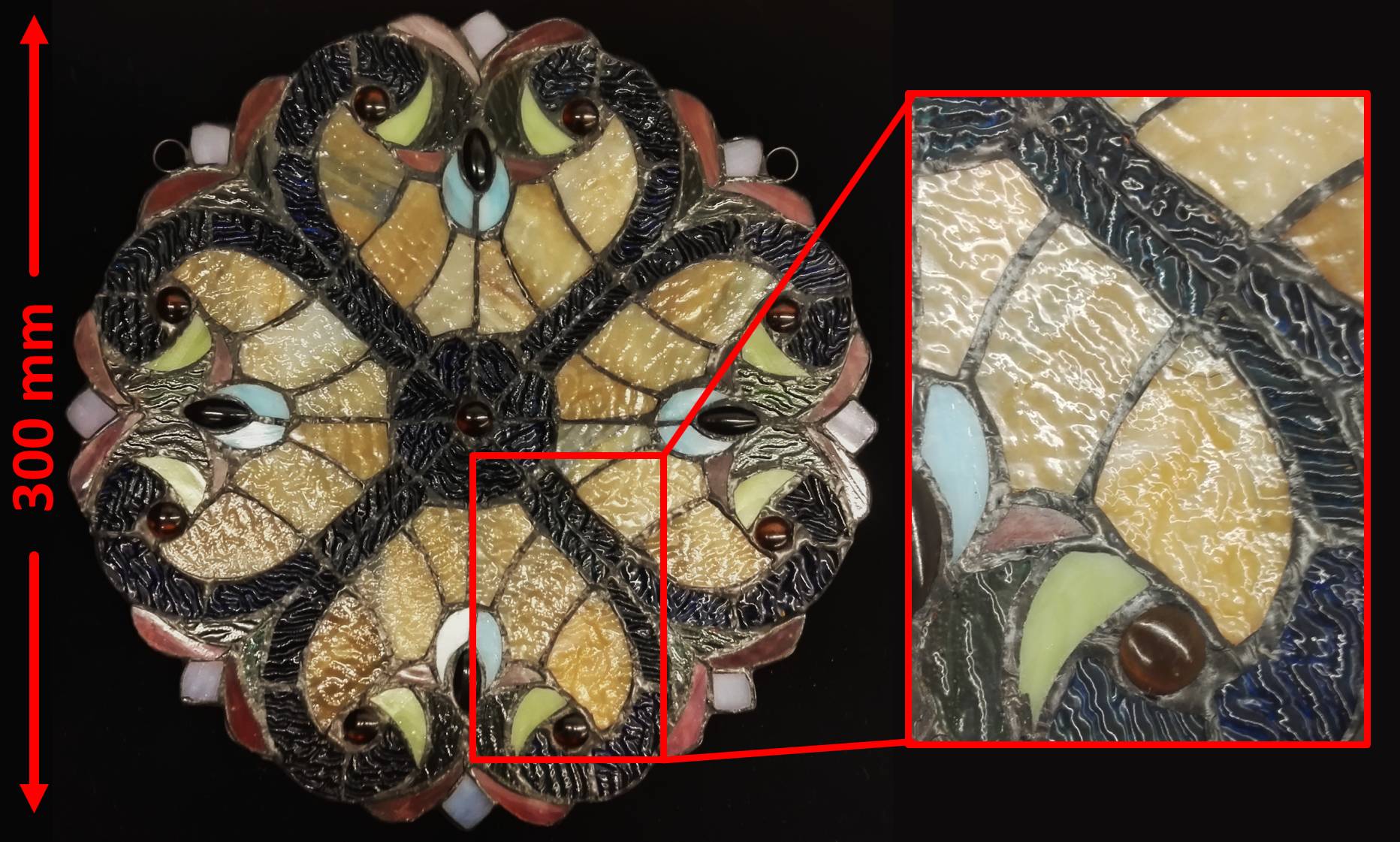}
\caption{Object to be measured with our system. One half of the glass painting is scanned by 14 views from different angles and  positions. }
\label{fig:GlassPainting}
\end{figure}

A single view measurement is not enough to capture an extended specular object with large normal variation in it's entirety. This is not only because of the limited effective field of view of mobile devices (see Fig.~\ref{fig:GradMaps}a), but also because the large normal variation of some surfaces cannot be captured from a single viewing angle (see e.g. Fig.~\ref{fig:NormalMap4Tiles}d). Our solution to this problem is to acquire and register multiple views of the object. In this section, we show qualitative results that demonstrate the feasibility of our approach. We study a circular shaped glass painting with a diameter of 300~mm. From the magnification window in Fig.~\ref{fig:GlassPainting}, it can be seen that the glass pieces in this painting exhibit high frequency surface features. Moreover, some glass pieces are milky, which introduces an additional challenge for the surface measurement. For the results shown below, we scanned one half of the glass painting by acquiring \textit{14 single views} under different viewing angles and positions.

\begin{figure}[t]
\centering \includegraphics[width= 0.9 \columnwidth]{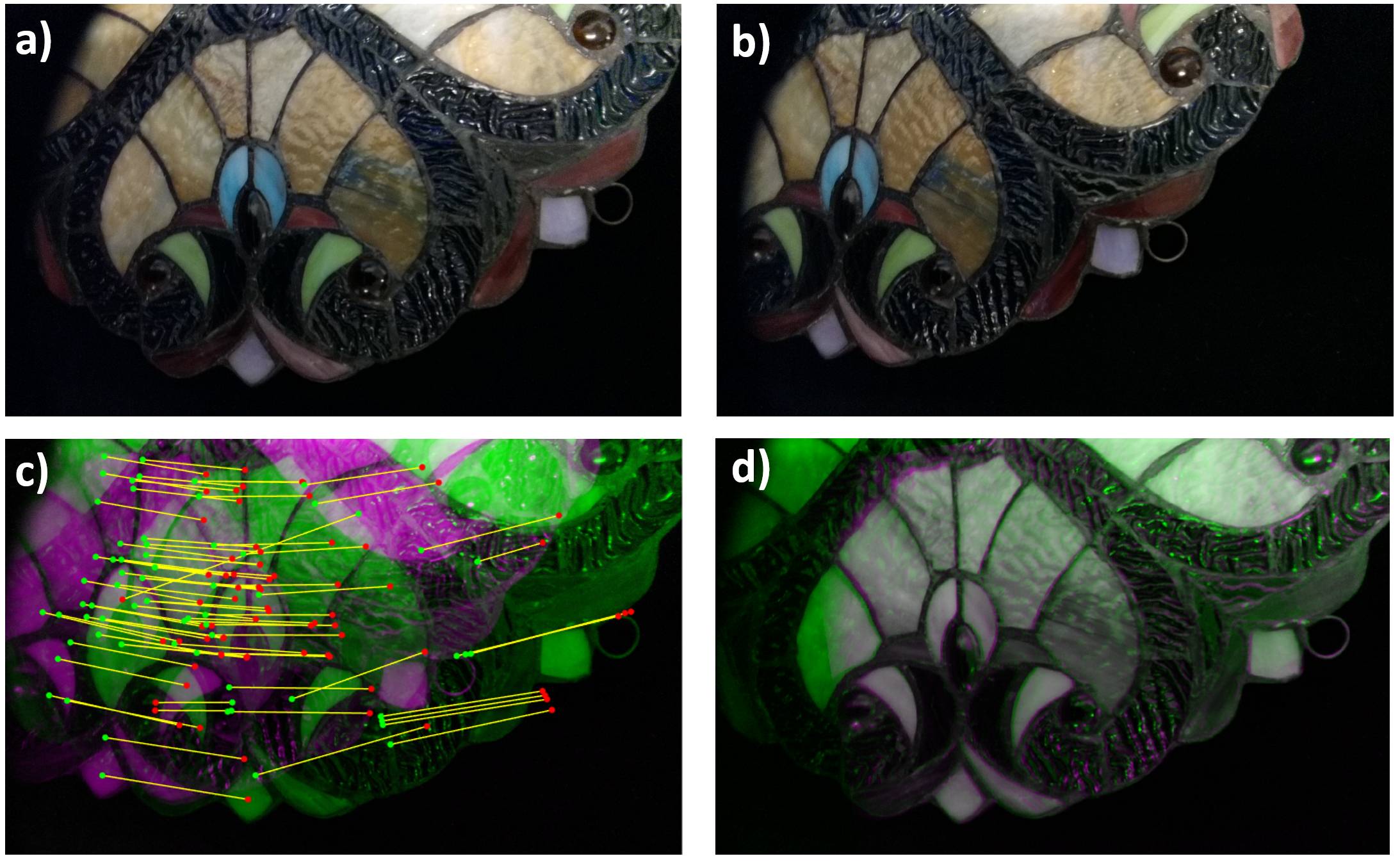}
\caption{Registration of two subsequent views. a) and b) `White images' (images captured) with black screen and diffuse room light  illumination) before distortion correction. c)~Detected and mapped features in the two subsequent `white images' (color-coded by green and magenta). d) Registered `white images'.}
\label{fig:RegExpl}
\end{figure}

To assist in registration, we acquired an additional `white image' (image of glass painting only illuminated by diffuse room light) at each viewing position. The registration transformation for the normal maps acquired at each single view is calculated from these `white images'. Performing registration with the `white images' was found to be more robust than registration with calculated normal maps. For registration, we used the feature based registration algorithms provided by the Matlab Computer Vision Toolbox. It should be noted that the usage of images which are captured under diffuse illumination is beneficial in this case, since the diffuse illumination makes the object look similar from different viewing angles. No strong specular reflections (which look different from different viewing angles) disturb the feature extraction of the registration algorithm. With this trick, we are able to register subsequent views without applying markers or other fiducials onto the object surface, just by using the texture of the object itself. Figure~\ref{fig:RegExpl} shows `white images' of two subsequent views, their detected and mapped features as well as the registration result.

It can be seen that the feature extraction and the subsequent registration transformation is applied on the \textit{whole} field of view of the camera (not only on the limited field in the middle) in order to detect a large number of features with high quality. In this case it is beneficial to perform a calibration of the front camera (e.g. with a checkerboard) to compensate for distortion. This reduces the registration error significantly. It should be also noted that such a distortion correction was avoided for the previous single-view measurements, since most of the signal was measured in the middle of the field of view, where the distortions are small. In the future, we hope to develop methods that estimate the distortion parameters of the camera \textit{during registration} without the need for an explicit calibration procedure.

\begin{figure}[t]
\includegraphics[width= \columnwidth]{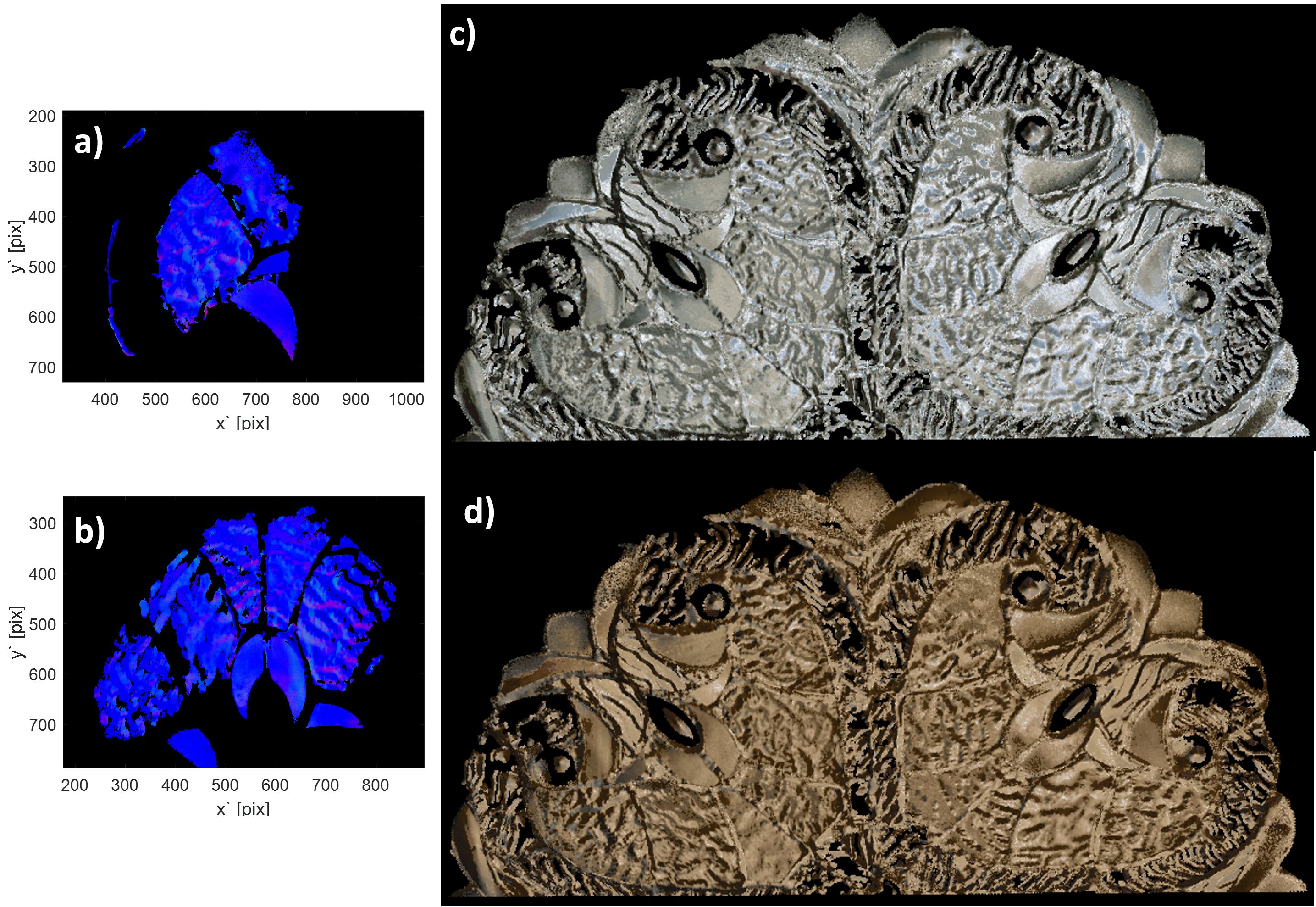}
\caption{Normal map measurement result for the multi-view measurement. The normal map consists of 14 single views which are stitched together via registration. a) and b) Two representative normal maps obtained from single views. c) and d) Normal map of the half glass painting under different shading. The map is stitched together from 14 registered single-views.}
\label{fig:RegResult}
\end{figure}

Our measurement results are displayed in Fig.~\ref{fig:RegResult}. Figures~\ref{fig:RegResult}a and b show two normal maps of the stained glass painting surface acquired from two different viewpoints. It can be seen that the 3D surface is not only recovered within the region in the center of the field of view. Depending on the orientation of the glass surface relative to camera and display, it is principally possible to detect normals within the whole field of view, though they might be sparse. Figures~\ref{fig:RegResult}c~and~d show all 14 views after registration and stitching under different shading. Most parts of the objects surface are densely reconstructed and the high frequency structures of the individual glass pieces are visible. However, some black regions are still present, mostly from the blue glass pieces in the painting. The structure of these pieces displays extraordinary high hills and deep craters, producing a wide distribution of normals that require more than 14 views to be measured effectively.

\section{Extension of the principle to different objects and surface types} 

Although the presented method was motivated by the 3D measurement of stained glass artworks, the system is in no case restricted to this specific object type. A 3D surface acquisition with our uncalibrated method is possible as long as the overall shape of the object is flat and the surface under test is relatively shiny.

\begin{figure}[t]
\centering \includegraphics[width= 0.8 \columnwidth]{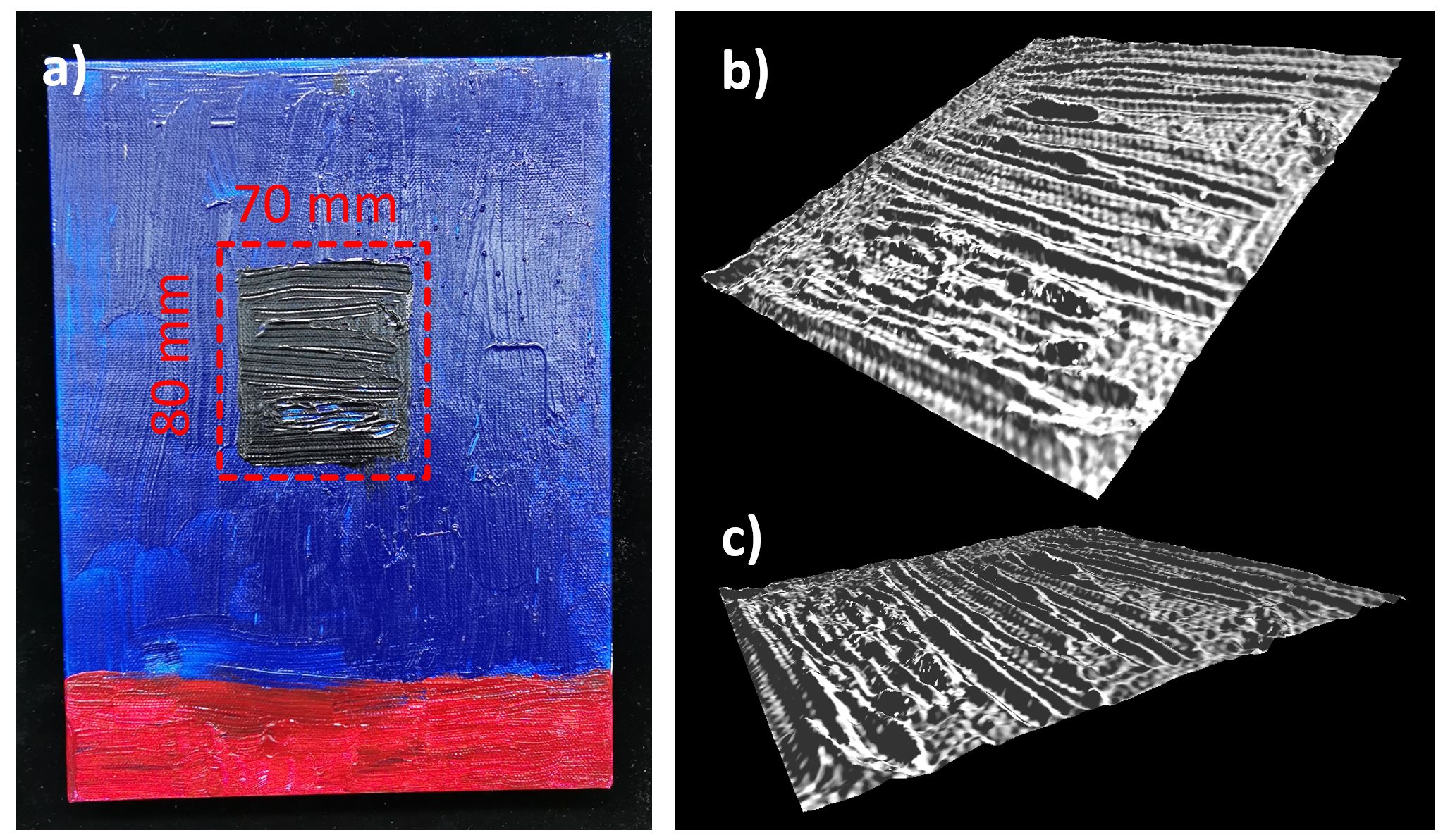}
\caption{Deflectometric measurement of a painting surface. a) Image of the painting. b)~and~c)~Surface shape of the marked region, calculated by integration of the acquired normal map. Brushstrokes an canvas can nicely be resolved.}
\label{fig:OilPainting}
\end{figure}

Figure~\ref{fig:OilPainting} displays the surface measurement of an oil painting. It should be noted that the three-dimensional analysis of painting surfaces is of great interest for the cultural heritage community. The ability to separate surface texture from its shape or slope data is an important tool for the analysis of painting techniques (e.g. by looking at the directions of brush strokes) and monitoring of pigment degradation in paintings \cite{Science, Smith}. Our mobile imaging method is well suited for the analysis of paintings 'in the wild', i.e. directly on the museum wall. Figure~\ref{fig:OilPainting}a shows an image of a measured oil painting. The surface normals of the black region in the red box (approximately $70mm \times 80mm$) are acquired with our method. For a better visualization of the hills and valleys of the brushstrokes, the acquired normal map is integrated to a depth map, using the Frankot-Chellappa surface integration algorithm~\cite{Frankot88}. Figures~\ref{fig:OilPainting}b and c show the calculated depth map from two different perspectives (z-component exaggerated for display purposes). Brush strokes and underlying canvas can nicely be resolved.

\begin{figure}[b]
\centering \includegraphics[width=  \columnwidth]{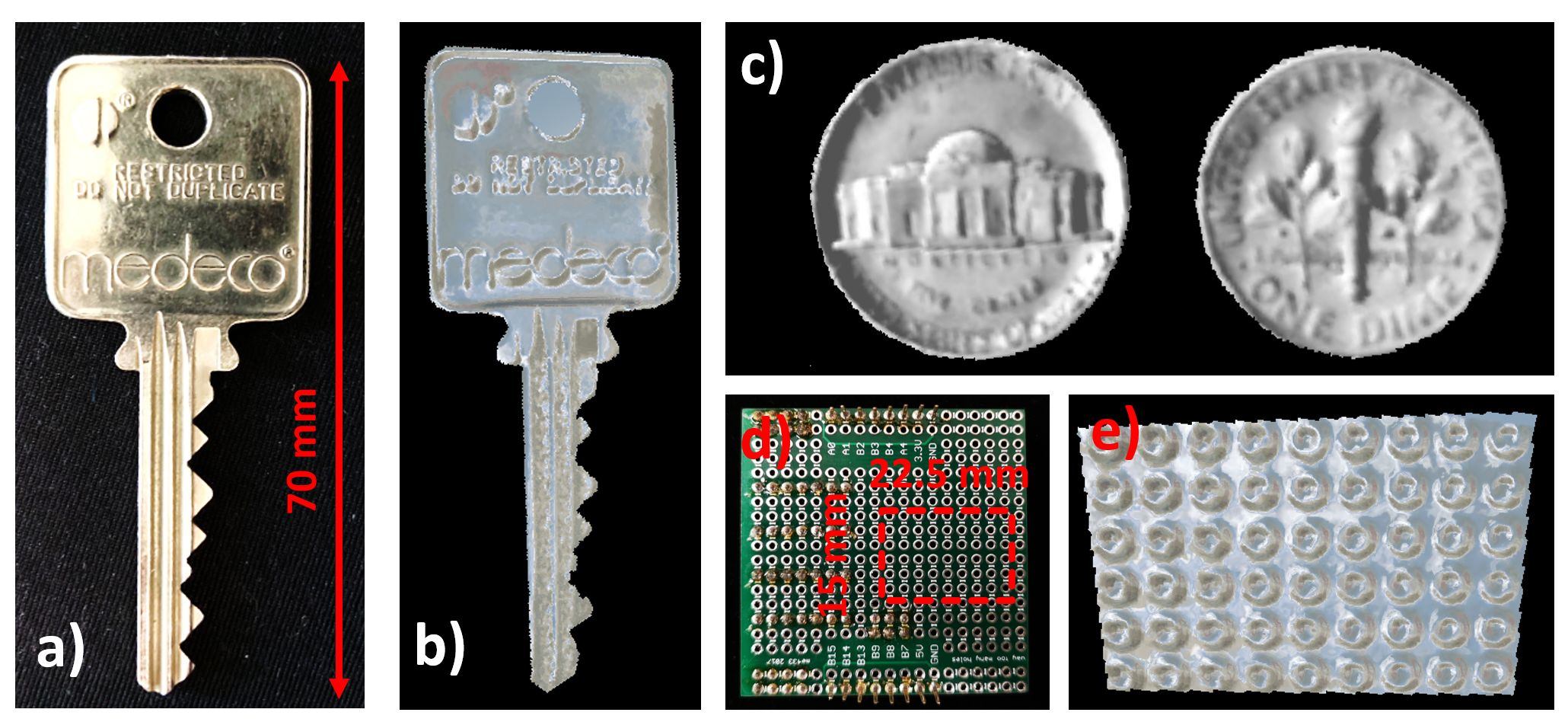}
\caption{Deflectometric measurements of technical / metallic surfaces. a) Image of measured key. b) Measured normal map of the key. c) Normal maps of a 5 cent and a 10 cent coin. d)~Image of measured circuit board. e) Measured normal map of the circuit board.}
\label{fig:KeyCoinBoard}
\end{figure}

Another potential field of application is the 3D acquisition of technical metallic surfaces. Measurement examples are shown in Fig.~\ref{fig:KeyCoinBoard}. Figure~\ref{fig:KeyCoinBoard}b displays the acquired normal map of a metallic key ($70 mm$ height, Fig.~\ref{fig:KeyCoinBoard}a), shaded with a specular finish. The normal maps of a 5~cent and a 10~cent US coin  ($21 mm$ / $18mm$ diameter) are shown in Fig.~\ref{fig:KeyCoinBoard}c. Imprinted letters or symbols can be resolved, both for the key as well as for the coins. Figure~\ref{fig:KeyCoinBoard}e displays the normal map of a circuit board (Fig.~\ref{fig:KeyCoinBoard}e)). The diameter of one single metallic ring is only about $2mm$. 

\begin{figure}[t]
\centering \includegraphics[width= 0.8\columnwidth]{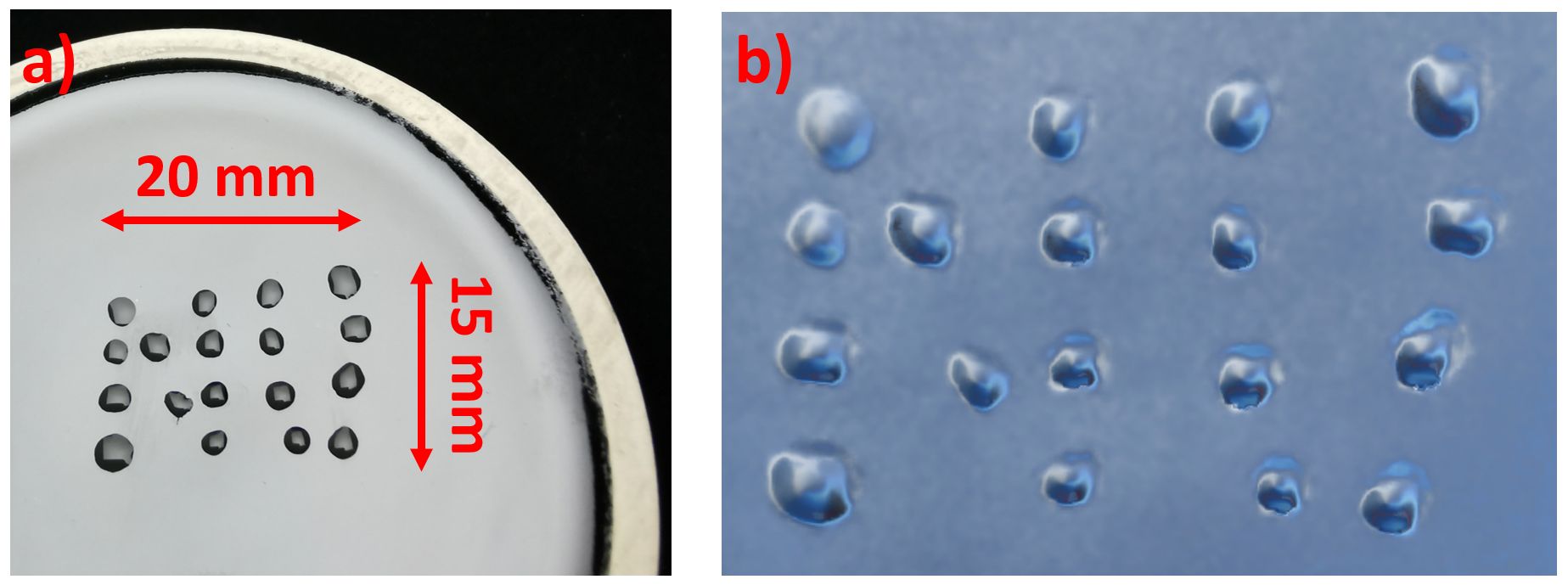}
\caption{Deflectometric measurement of fluid surfaces. a) Water drops on an enameled ceramic surface (coffee mug). b) Evaluated normal map.}
\label{fig:Waterdrops}
\end{figure}

In a last example, we demonstrate the capability of our method to measure fluid surfaces, e.g. for the analysis of surface tension. Figure~\ref{fig:Waterdrops}b shows the normal map that was acquired from water drops on an enameled ceramic surface (coffee mug). The water drops are arranged to form the letters `N U' (Fig.~\ref{fig:Waterdrops}a). The shape of each drop is clearly visible from the normal map. In the future, we plan to use our system to implement a single shot PMD technique~\cite{LiuSS} to measure \textit{dynamic fluid surfaces}. In addition, we are developing algorithms capable of recovering surface normals from objects with much more complicated reflectivity


\section{Conclusion and Outlook}  \label{sec:ConcOut}

In this paper we presented a  method and system that is able to perform Deflectometry measurements of  specular surfaces with a mobile device. The intended application of our system is the three dimensional slope measurement of extended specular surfaces. Stained glass artworks are very compelling objects to demonstrate the power of our measurement technique. Though the overall surface shape of these objects is relatively flat, the surfaces display high frequency features and a large variance in surface normals. We demonstrated experimentally some of the challenges in measureming these 3D surfaces using the small screen of a mobile device. In order to sample the entire object surface densely with high resolution over a large field of view, we applied a feature based registration to stitch normal maps from different viewing angles and positions.

We demonstrated the 3D surface measurement of stained glass surfaces using both single view and registered multi-view measurements. Our results provide access to 3D surface information which can be used as a fingerprint for artworks, and can help establish authenticity of an artwork. As a proof of principle, we scanned one half of a circular stained glass artwork with 300~mm diameter by stitching together \textit{14 single views}. In a second experiment not shown in the paper, we tried registration with 28 views. However, global registration errors were significant so that the first and last views did not fit together after one pass. This is a well known problem for surface measurements with registration \cite{2014arXiv1401.1946A}. Reducing the global registration error is one of our main goals for future work.

We demonstrated first steps toward a self-calibrating Deflectometry system.
Our single-view results exploited a-priori knowledge about the object to avoid extensive fringe and display calibration, and avoiding the depth-normal ambiguity problem without use of a second camera \cite{doi:10.1117/12.545704}. In the future, we hope to develop self-calibrating algorithms for multi-view measurements. Our plan is to apply a non-rigid registration on our data and obtain the information about the distortion from the calculated deformation fields. Moreover, we will work towards obtaining \textit{quantitative measurements without calibration}. This work will build upon previously demonstrated self-calibrating PMD setups, e.g. in~\cite{Olesch2012DeflectometricSF}. 

Although we have shown that hand-guided measurements are possible with our system, PMD is commonly a multi-shot principle, and can therefore introduce motion blur. Single-shot PMD techniques that rely on single-sideband demodulation, e.g., like introduced in ~\cite{LiuSS} will not work on  objects with high surface frequencies because of the severe bandwidth restrictions. In the future we want to explore other single-shot Deflectometry techniques, e.g. displaying a binary line pattern on the screen. Such 'binary' single-shot techniques do not acquire dense 3D surface information, and therefore require more sophisticated registration algorithms. Examples of how such problems are solved in the field of line triangulation can be found in~\cite{2014arXiv1401.1946A, doi:10.1063/1.4809687, Willomitzer:17}. Our future goal is to develop similar methods for Deflectometry. Ideally, the user need only continuously wave around his device in front of the object to obtain a dense 3D reconstruction after only a few seconds.

Lastly, to foster adoption of our technique by a broad audience, we plan to make our measurement App publicly available so that anyone with a mobile device can make 3D surface measurements of specular objects. Each user will be able to transform his phone into a 3D measurement instrument. Captured images will be streamed to our servers, where 3D surface information is calculated, stored for further use, and presented back to the user. We hope this framework will serve as a platform for crowd-sourced aggregation of surface shape acquisition/fingerprinting of unattributed artworks around the globe.  

\section*{Acknowledgments}
The authors thank G. H\"ausler and C. Faber for the fruitful discussions.


\bibliography{sample}






\end{document}